# PCG-Cut: Graph Driven Segmentation of the Prostate Central Gland

**Jan Egger***

Department of Medicine, University Hospital of Marburg (UKGM), Marburg, Hesse, Germany

## Abstract

Prostate cancer is the most abundant cancer in men, with over 200,000 expected new cases and around 28,000 deaths in 2012 in the US alone. In this study, the segmentation results for the prostate central gland (PCG) in MR scans are presented. The aim of this research study is to apply a graph-based algorithm to automated segmentation (i.e. delineation) of organ limits for the prostate central gland. The ultimate goal is to apply automated segmentation approach to facilitate efficient MR-guided biopsy and radiation treatment planning. The automated segmentation algorithm used is graph-driven based on a spherical template. Therefore, rays are sent through the surface points of a polyhedron to sample the graph's nodes. After graph construction – which only requires the center of the polyhedron defined by the user and located inside the prostate center gland – the minimal cost closed set on the graph is computed via a polynomial time s-t-cut, which results in the segmentation of the prostate center gland's boundaries and volume. The algorithm has been realized as a C++ module within the medical research platform MeVisLab and the ground truth of the central gland boundaries were manually extracted by clinical experts (interventional radiologists) with several years of experience in prostate treatment. For evaluation the automated segmentations of the proposed scheme have been compared with the manual segmentations, yielding an average Dice Similarity Coefficient (DSC) of 78.94±10.85%.

**Citation:** Egger J (2013) PCG-Cut: Graph Driven Segmentation of the Prostate Central Gland. PLoS ONE 8(10): e76645. doi:10.1371/journal.pone.0076645

**Editor:** Zoran Culig, Innsbruck Medical University, Austria

**Received** July 25, 2013; **Accepted** August 30, 2013; **Published** October 11, 2013

**Copyright:** © 2013 Jan Egger. This is an open-access article distributed under the terms of the Creative Commons Attribution License, which permits unrestricted use, distribution, and reproduction in any medium, provided the original author and source are credited.

**Funding:** This author has no support or funding to report.

**Competing Interests:** The author has no potential conflict of interests.

* E-mail: egger@med.uni-marburg.de

## Introduction

Prostate cancer is the most abundant cancer in men, with over 200,000 expected new cases and around 28,000 deaths in 2012 in the US alone [1]. With many prostate cancers being of low aggressiveness and a high complication rate of radical prostatectomy (impotence, incontinence), accurate risk stratification for each individual cancer is central to a successful treatment strategy. To date prostate cancer diagnosis is widely based on prostate specific antigen (PSA) level testing and transrectal ultrasound (TRUS) guided biopsies, associated with a low specificity (PSA testing) or a low sensitivity (TRUS biopsies) resulting in high rates of rebiopsies. Diagnostic prostate magnetic resonance imaging (MRI) and MRI guided prostate biopsies were introduced clinically to resolve the shortcomings of the aforementioned methods, improving diagnostic discrimination rates [2]. The goal of this work is to enhance the state of the art in automated segmentation (i.e. delineation) of organ limits for the prostate, a step that has been shown to facilitate efficient MR-guided biopsy and radiation treatment planning.

Others working in the area of prostate segmentation are Ghose et al. [3], which introduced a graph cut based energy minimization [4] of the posterior probabilities obtained in a supervised learning schema for automatic 3D segmentation of the prostate in MRI data. Thus, the probabilistic classification of the prostate voxels is achieved with a probabilistic atlas and a random forest based learning framework. Furthermore, the posterior probabilities are combined to obtain the likelihood of a voxel belonging to the prostate and afterwards the 3D graph cut based energy minimization in the stochastic space provides segmentation of the prostate. Ghose et al. [5] also recently presented a comprehensive survey of prostate segmentation methodologies in ultrasound (US), magnetic resonance (MR) and computed tomography (CT) images. Amongst others, they discuss edge based [6–8], atlas based [9–11] and hybrid methods [12–14] for Prostate segmentation in MR images. For a detailed description of these approaches the reader is referred to the survey of Ghose. However, in the meantime there has been a prostate MR image segmentation challenge from the MICCAI society [15] and these (additional) approaches are presented in more detail in this section. Vincent et al. [16] introduce a fully automatic segmentation of the prostate using Active Appearance Models (AAM) [17]. Therefore, high quality correspondences for the model are generated using a Minimum Description Length (MDL) Groupwise Image Registration method [18] and a multi start optimisation scheme was used to robustly match the model to new images. Birkbecky et al. [19] present a region-specific hierarchical segmentation of MR Prostate images by using discriminative learning. After normalizing intra- and inter-image intensity variation, they used Marginal Space Learning (MSL) [20] to align a statistical mesh model on the image. Thus, the mesh is hierarchically refined to the image boundary using spatially varying surface classifiers. Malmberg et al. [21] introduce Smart Paint, which is an interactive segmentation method applied to MR prostate segmentation. Thereby, the user interaction is inspired by the way an airbrush is used and objects are segmented by sweeping with the mouse cursor in the image. The proposed segmentation tool allows the user to add or remove

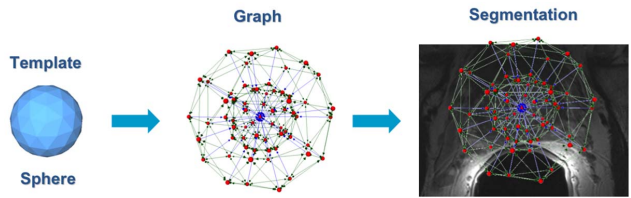

**Figure 1. Principle of the Nugget-Cut Scheme: A spherical template (left) is used as a basic structure for the segmentation graph (middle), which is created inside the image (right).**
doi:10.1371/journal.pone.0076645.g001

details in 3D and the user interface shows the segmentation result in 2D slices through the object. Ou et al. [22] present in their contribution a multi-atlas-based automatic pipeline for segmenting prostate in MR images. In a first step they register all atlases onto the target image to obtain an initial segmentation of the prostate. Afterwards, they *zoom-in* which means they re-run all atlas-to-target registrations. However, this time they restrict the registration to the vicinity of the prostate, ignoring compounding structures that are far away from the prostate and are largely variable. Kirschner et al. [23] used a probabilistic Active Shape Model (ASM) [24] and [25] for automatic prostate segmentation in MR images. However, first they employ a boosted prostate detector to locate the prostate in the images. Thus, they extend the Viola-Jones object detection algorithm [26] to 3D and afterwards they use the probabilistic ASM (PASM) for the delineation of the prostate contour. A convex optimization approach for 3D Prostate MRI Segmentation with generic Star shape prior has been proposed by Yuan et al. [27]. Thereby, the approach incorporates histogram-matching and a variational formulation of a generic star shape prior [28] and [29]. The presented generic star shape prior provides robustness to the segmentation when the images suffer from poor quality, noise, and artifacts. Gao et al. [30] introduce an automatic multi-atlas based prostate segmentation using local appearance-specific atlases and patch-based voxel weighting. Therefore, the atlases with the most similar global appearance are classified into the same categories and the sum-of-square local intensity difference after affine registration is used for an atlas selection. After the non-rigid registration, a local patch-based atlas fusion is performed using voxel weighting based on the local patch distance. Another contribution that uses Active Appearance

Models in 3D for prostate MR image segmentation is presented by Maan and van der Heijden [31]. In a first step, the shape context based non-rigid surface registration of the manual segmented images was used to obtain the point correspondence between the given training cases. Thereby, their contribution builds on the work of Kroon et al. [32] where knee cartilages have been segmented. In the second step, the AAM was used to segment the prostate on 50 training cases. Litjens et al. [33] introduced a multi-atlas approach for prostate segmentation in MR images, where the atlases are registered using localized mutual information as a metric. Afterwards, the Selective and Iterative Method for Performance Level Estimation (SIMPLE)-algorithm [11] was used to merge the atlas labels and obtain the final segmentation. Ghose et al. [34] proposed a Random Forest based [35] classification approach for prostate segmentation in MR scans. Thus, they introduce a supervised learning framework of decision forest to achieve a probabilistic representation of the prostate voxels. Then, propagation of region based level-sets in the stochastic space provides [36] and [37] the segmentation of the prostate. Toth and Madabhushi [38] use deformable landmark-free Active Appearance Models to segment prostate MRI data. Therefore, a deformable registration framework was created to register a new image to the trained appearance model, which was subsequently applied to the prostate shape to yield a final segmentation. Yin et al. [39] proposed a fully automated 3D prostate central gland segmentation in MR images. Thus, they applied the Layered Optimal Graph Image Segmentation of Multiple Objects and Surfaces (LOGISMOS) approach. The LOGISMOS model contained both: shape and topology information during deformation and they generated the graph cost by training classifiers and they used coarse-to-fine search. Moreover, the authors want to point the reader at this point to a book chapter that includes a section about segmentation for prostate interventions [40].

The purpose of this contribution is to introduce the results of a graph based segmentation algorithm for the prostate central gland (PCG) in MR scans. The algorithm uses a spherical template and sends rays through the surface points of a polyhedron to sample the graph's nodes. Even if there exist graph-based segmentation methods in 2D for MR prostate images and some hybrid 3D methods for MR prostate images (like [41] and [42]), the author is not aware of a *pure* 3D graph-based method that has been applied to the PCG in MRI scans. For evaluation the Dice Similarity Coefficient (DSC) [43] and [44] a common evaluation metrics in medical image processing was applied, which has also been used in the most of the existing contributions.

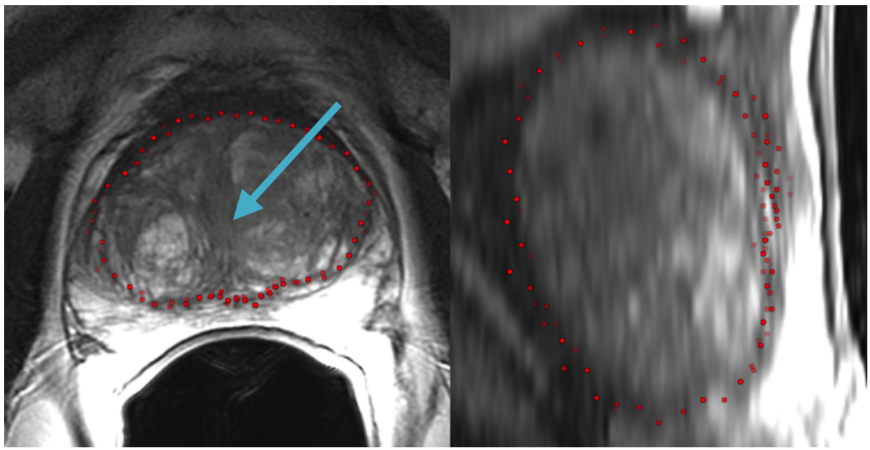

**Figure 2. Segmentation results (red) in axial and sagittal reformatting (blue arrow: position of the seed point).**
doi:10.1371/journal.pone.0076645.g002

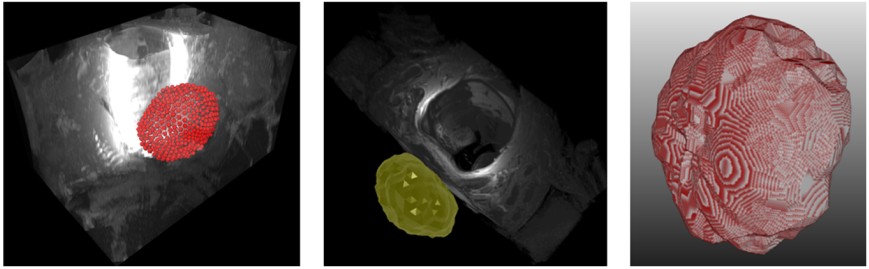

**Figure 3. 3D visualizations of segmentation results: segmentation nodes and triangulated segmentation result with surrounding structures (left and middle), and voxelized mask of the segmented prostate (right).**
doi:10.1371/journal.pone.0076645.g003

The contribution is organized as follows. Section 2 presents the materials and the methods. Section 3 presents the results of the experiments, and Section 4 concludes and discusses the contribution and outlines areas for future work.

## Materials and Methods

### Data

For testing the approach T2-weighted magnetic resonance imaging datasets from the clinical routine have been used. The datasets applied in the evaluation are freely available in the internet and can be used for research purposes [45]: http://www.na-mic.org/publications/item/view/2174.

### Algorithm

The Nugget-Cut scheme [46] was applied for prostate center gland segmentation. It sets up a directed 3D-graph G(V,E) in two steps: (I) sending rays through the surface points of a polyhedron [47] and (II) sampling the graph's nodes n∈V along every ray. Additionally, a corresponding set of edges e∈E is generated, which consists of edges between the nodes and edges that connect the nodes to a source s and a sink t. After graph construction – the center of the polyhedron was defined by the user and located inside the prostate center gland – the minimal cost closed set on the graph is computed via a polynomial time s-t-cut [48], which results in the segmentation of the prostate center gland's boundaries and volume. The overall principle of the Nugget-Cut scheme that has been applied to the PCG segmentation is presented in Figure 1. As shown in the figure, a spherical template (left) is used as a basic structure for setting up the segmentation graph (middle). Finally, this graph is generated inside the image with its center at the position of the user-defined seed point (right).

By definition the edges/arcs $<v_i, v_j> \in E$ of the graph $G$ connect two nodes $v_i, v_j$ and there exist two types of ∞-weighted arcs: z-arcs $A_z$ and r-arcs $A_r$, whereby $Z$ is the number of sampled points along one ray $z = (0,...,Z-1)$ and $R$ is the number of rays sent out to the surface points of a polyhedron $r = (0,...,R-1)$. $V(x_n,y_n,z_n)$ is one neighbor of $V(x,y,z)$, or in other words $V(x_n,y_n,z_n)$ and $V(x,y,z)$ belong to the same triangle in case of a triangulation of the polyhedron:

$$A_z = \{\langle V(x,y,z), V(x,y,z-1)\rangle | z > 0\}$$

$$A_r = \{\langle V(x,y,z), V(x_n,y_n, \max(0, z-\Delta_r))\rangle\}$$

The ∞-weighted arcs between two nodes along a ray $A_z$ ensure that all nodes below the polyhedron surface in the graph are included to form a closed set. According to this, the interior of the object is separated from the exterior in the data. On the other hand, arcs $A_r$ between the nodes of different rays constrain the set of possible segmentations and enforce smoothness via a parameter $\Delta_r$ and the larger this parameter is, the larger is the number of possible segmentations. Finally, the s-t cut creates an optimal segmentation of the PCG under influence of the parameter $\Delta_r$ that controls the stiffness of the boundaries. For example, a $\Delta$ value of zero ($\Delta_r = 0$) ensures that the segmentation result is a sphere and the position of the sphere within the image is based on the edges to the source and sink (s-t-edges). The weights $w(x, y, z)$ for every s-t-edge are assigned in the following manner: weights are set to $c(x,y,z)$ when $z$ is zero and otherwise to $c(x,y,z)−c(x,y,z-1)$. In doing so, $c(x,y,z)$ is the absolute value of the intensity difference between an average grey value of the PCG and the grey value of the voxel at position $(x,y,z)$. Note that the average grey value of the PCG can automatically be estimated around the user-defined seed point inside the Prostate.

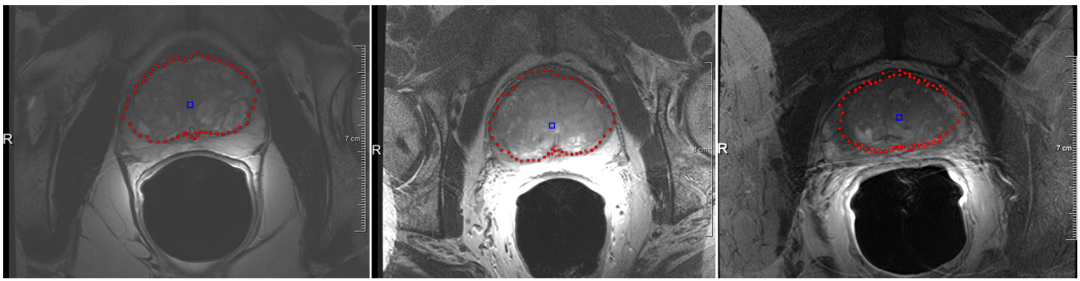

**Figure 4. Segmentation results (red) for three different cases of proposed algorithm at the _axial_ height of the user-defined seed point (blue).**
doi:10.1371/journal.pone.0076645.g004

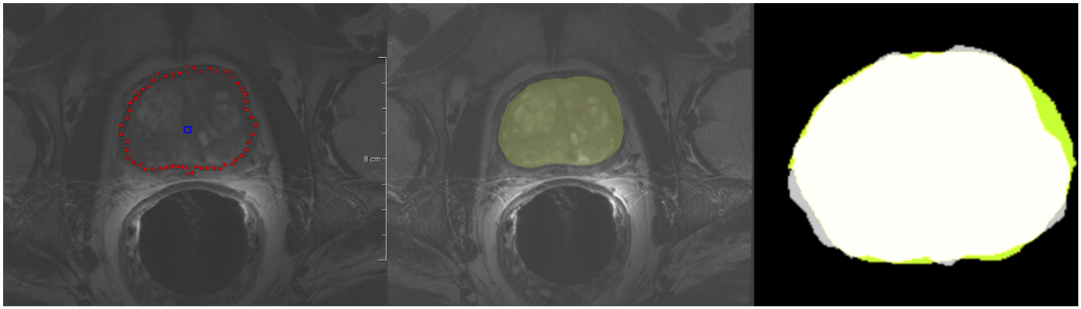

**Figure 5. Direct comparison of automatic (red) and manual (yellow) segmentation results for the same case: automatic segmentation result (left), manual segmentation (middle) and superimposed visualization of both segmentations (right).**
doi:10.1371/journal.pone.0076645.g005

## Results

A C++ module was implemented within the medical prototyping platform MeVisLab (see http://www.mevislab.de) for evaluation. In the C++ implementation the overall segmentation: (1) sending rays, (2) graph construction and (3) mincut computation, took about one second on a Macbook Pro laptop computer with an Intel Core i7-2860QM CPU, 4×2.50 GHz, 8 GB RAM, Windows 7 Professional ×64.

Figure 2 presents screenshots of an automated segmentation result (red) in axial (left) and sagittal reformatting (right) slices. The blue arrow in the left image points to the position of the user-defined seed point. Figure 3 shows several 3D visualizations of segmentation results of the PCG. On the left side of the segmentation nodes (red) of the mincut are presented. The image in the middle shows the corresponding triangulated segmentation result (green/yellow) with surrounding anatomical structures. Finally, the rightmost image presents the voxelized mask of the segmented prostate, which has been used to calculate the Dice Similarity Coefficient with the manual slice-by-slice segmented prostate. Segmentation results (red) for three different cases of the proposed algorithm at the *axial* height of the user-defined seed point (blue) are displayed in Figure 4. A direct comparison of an automated (red) and a manual (yellow) segmentation for the same case is presented in Figure 5. The automated segmentation result is shown on the left side of the figure, the manual segmentation in the middle and the superimposed visualization of both segmentations on the right. Finally, Figure 6 displays several axial slices containing the automated segmentation results (red) for a case. The fourth slice from the left includes also the user-defined seed point from which the segmentation graph has been created.

Table 1 presents the results for a direct comparison of manual slice-by-slice and PCG-Cut segmentation results for ten prostate central glands via the Dice Similarity Coefficient. Table 2 presents the summary of the results from Table 1, including the minimum (min), the maximum (max), mean $\mu$ and standard deviation $\sigma$ for ten prostate central glands.

## Conclusion and Discussion

In this study, the segmentation results for the prostate gland in MRI data using a recently developed method have been presented. Therefore, a graph driven method has been applied that bases on a spherical template. The algorithm prefers spherically- and elliptically-shaped 3D objects and has already been evaluated with glioblastoma multiforme (GBM), pituitary adenoma and cerebral aneurysm data [49]. To sample the graph's nodes rays are sent through the surface points of a polyhedron and afterwards the minimal cost closed set on the graph is computed via a polynomial time s-t-cut, which results in the segmentation of the prostate central gland boundaries and volume. Thereby, the

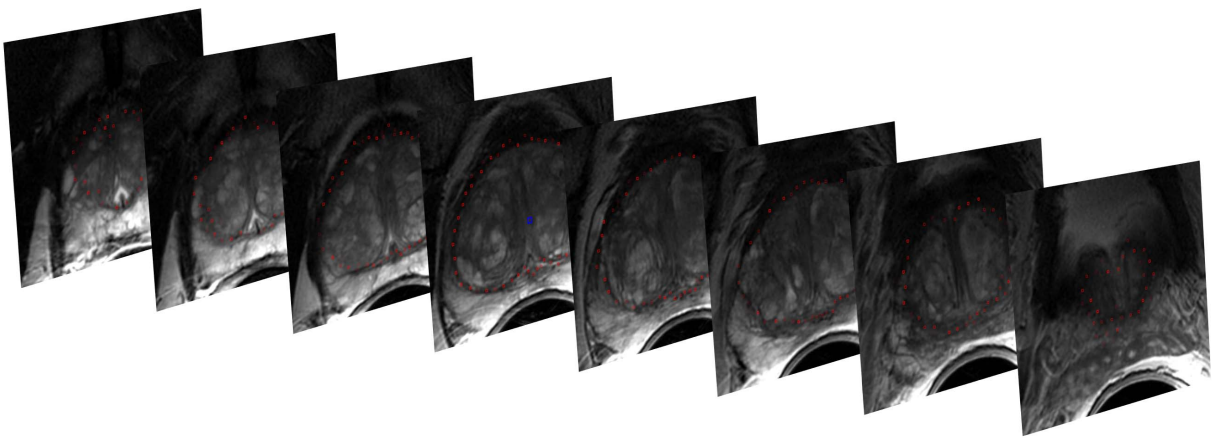

**Figure 6. Several axial slices with automatic segmentation results (red).** The fourth slice from the left contains also the user-defined seed point from which the segmentation graph has been created.
doi:10.1371/journal.pone.0076645.g006

**Table 1.** Direct comparison of manual slice-by-slice and PCG-Cut segmentation results for ten prostate central glands (PCG) via the Dice Similarity Coefficient (DSC).

| Case No. | volume of PCG (mm³) | | number of voxels | | DSC (%) |
|---|---|---|---|---|---|
| | manual | Automatic | manual | automatic | |
| 1 | 20820.8 | 41692.1 | 682256 | 1366168 | 61.79 |
| 2 | 13670.3 | 16487.9 | 447949 | 540277 | 62.57 |
| 3 | 29006 | 31706.7 | 1419342 | 1551511 | 84.79 |
| 4 | 51418.2 | 56490.6 | 1684871 | 1851084 | 88.79 |
| 5 | 44258.4 | 40306 | 2165709 | 1972307 | 89.42 |
| 6 | 66161.6 | 67559.2 | 2167986 | 2213781 | 88.76 |
| 7 | 15317.8 | 13312.5 | 501932 | 436224 | 83.93 |
| 8 | 32826.1 | 38079.4 | 1075646 | 1247786 | 75 |
| 9 | 22047.6 | 13289.7 | 722456 | 435477 | 69.68 |
| 10 | 17718.1 | 16886 | 866648 | 825950 | 84.71 |

doi:10.1371/journal.pone.0076645.t001

**Table 2.** Summary of results: min, max, mean $\mu$ and standard deviation $\sigma$ for ten prostate central glands (PCG).

| | volume of PCG (cm³) | | number of voxels | | DSC (%) |
|---|---|---|---|---|---|
| | manual | automatic | manual | automatic | |
| min | 13.67 | 13.29 | 447949 | 435477 | 61.79 |
| max | 66.16 | 67.56 | 2167986 | 2213781 | 89.42 |
| $\mu \pm \sigma$ | 31.32±17.45 | 33.58±18.88 | 1173479.5 | 1244056.5 | 78.94±10.85 |

doi:10.1371/journal.pone.0076645.t002

biopsy and radiation treatment planning during an intervention. And due to the specific graph construction, the segmentation result for the introduced approach can be calculated within one second (including graph construction and mincut calculation). Thus, a repositioning of the seed point and a recalculation of the segmentation result (in case of an unsatisfied segmentation) can be performed very fast, therefore making the approach eligible intraoperative MR-guidance (in general, a few replacements of the seed point had to be done for every case to achieve the presented results). Furthermore, most existing semi-automatic approaches need a more time-consuming initialization. Often areas inside and outside of the structure (e.g. the prostate) have to be defined by a user before the segmentation can be started.

There are several areas of future work, including a direct comparison with other approaches on the same datasets and extensions of the automated segmentation to structures adjacent to the central zone of the prostate gland, such as the peripheral prostatic zone (Prostate-Cut). An immediate application for MR-guided biopsy is the generation of regions of interest as an aid to the automated registration of preoperative to intraprocedural images [44]. Moreover, the integration of a manual refinement method into the automatic algorithm [50] and [51] is planned and providing the approach as a module for 3D Slicer (http://www.slicer.org/) [52] to the community.

approach requires only the center of the polyhedron defined by the user and located inside the prostate center gland and the algorithm has been realized as a C++ module within the medical research platform MeVisLab. As the reference for comparison, central gland boundaries manually extracted by interventional radiologists with several years of experience in prostate imaging have been used (see the details in Fedorov et al. [44]). Then, the segmentation results obtained using the proposed scheme have been compared with the manual segmentations, yielding an average Dice Similarity Coefficient around 80%. In summary, the research highlights are:

- A graph-based approach has been developed and applied to automatic PCG segmentation.
- Manual slice-by-slice segmentations of prostate central glands (PCG) have been performed by clinical experts resulting in ground truth of PCG boundaries.
- The quality of the automated segmentations have been evaluated with the Dice Similarity Coefficient.

The most existing approaches from the literature aim to automatic or even fully automatic process large amounts of prostate datasets to support time consuming manual slice-by-slice segmentations. In contrast, this is in general not possible with the presented approach because the user still needs to define a seed point for the graph inside the prostate central glad. In fact, the aim of the presented approach is to facilitate efficient MR-guided

## Acknowledgments

In the first place, the author would like to thank *Fraunhofer MeVis* in Bremen, Germany, for their collaboration and especially Horst K. Hahn for his support. Furthermore, the author thanks Drs. Fedorov, Tuncali, Fennessy and Tempany for sharing the prostate data collection and Dr. med. Tobias Penzkofer for his feedback and proofreading.

## Author Contributions

Conceived and designed the experiments: JE. Performed the experiments: JE. Analyzed the data: JE. Contributed reagents/materials/analysis tools: JE. Wrote the paper: JE.